\documentclass[conference]{IEEEtran}
\IEEEoverridecommandlockouts
\usepackage{cite}
\usepackage{amsmath,amssymb,amsfonts}
\usepackage{algorithmic}
\usepackage{graphicx}
\usepackage{textcomp}
\usepackage{xcolor}

\usepackage{algorithm}
\usepackage{array}
\usepackage{tabularx}
\usepackage{booktabs}
\usepackage{multirow}
\usepackage{colortbl}
\usepackage{hhline}
\def\BibTeX{{\rm B\kern-.05em{\sc i\kern-.025em b}\kern-.08em
    T\kern-.1667em\lower.7ex\hbox{E}\kern-.125emX}}

\begin{document}

\title{Dense Point Clouds Matter: Dust-GS for Scene Reconstruction from Sparse Viewpoints}



\author{
    \IEEEauthorblockN{
        Shan Chen\IEEEauthorrefmark{1}, 
        Jiale Zhou\IEEEauthorrefmark{1}, 
        Lei Li\IEEEauthorrefmark{2}\IEEEauthorrefmark{3}
    }
    \IEEEauthorblockA{
        \IEEEauthorrefmark{1}East China University of Science and Technology\\
        \IEEEauthorrefmark{2}University of Washington 
        \IEEEauthorrefmark{3}University of Copenhagen\\
    }
}

\maketitle

\begin{abstract}

3D Gaussian Splatting (3DGS) has demonstrated remarkable performance in scene synthesis and novel view synthesis tasks. Typically, the initialization of 3D Gaussian primitives relies on point clouds derived from Structure-from-Motion (SfM) methods. However, in scenarios requiring scene reconstruction from sparse viewpoints, the effectiveness of 3DGS is significantly constrained by the quality of these initial point clouds and the limited number of input images. In this study, we present Dust-GS, a novel framework specifically designed to overcome the limitations of 3DGS in sparse viewpoint conditions. Instead of relying solely on SfM, Dust-GS introduces an innovative point cloud initialization technique that remains effective even with sparse input data. Our approach leverages a hybrid strategy that integrates an adaptive depth-based masking technique, thereby enhancing the accuracy and detail of reconstructed scenes. Extensive experiments conducted on several benchmark datasets demonstrate that Dust-GS surpasses traditional 3DGS methods in scenarios with sparse viewpoints, achieving superior scene reconstruction quality with a reduced number of input images. 
\end{abstract}


\section{Introduction}

The synthesis of high-quality 3D scenes from sparse image data has emerged as a crucial research area in computer vision and graphics, driven by applications such as virtual reality, augmented reality, autonomous driving, and robotics \cite{tonderski2024neurad,pan2023uniocc,zhou2024drivinggaussian}. These applications require accurate 3D reconstructions from limited viewpoints, often with only a few input images available. While traditional methods like Neural Radiance Fields (NeRF) \cite{martin2021nerf} have shown impressive performance in generating detailed 3D scenes and novel views, they rely on implicit volumetric representations that demand substantial computational resources for pixel-level ray tracing and optimization \cite{mildenhall2021nerf,wang2023sparsenerf,truong2023sparf}. This dependency makes them less practical for scenarios with sparse input data where computational efficiency and rapid scene reconstruction are essential.

Recently, 3D Gaussian Splatting (3DGS) \cite{kerbl20233d,li2024dngaussian,lu2024scaffold} has emerged as a promising alternative, offering explicit representations for 3D scene synthesis that bypass the computational burden of voxel-based rendering of NeRF. By utilizing Gaussian primitives, 3DGS facilitates faster and more efficient scene reconstruction and novel view synthesis \cite{li2024dngaussian,zhang2024pixel,yu2024mip}. The explicit nature of Gaussian representations allows for direct sampling and manipulation in space, enhancing both rendering efficiency and flexibility in editing 3D scenes. However, the effectiveness of 3DGS is inherently tied to the quality of the initial point clouds generated by Structure-from-Motion (SfM) \cite{schoenberger2016mvs,schonberger2016structure} methods. In scenarios with sparse viewpoints and limited input images, the initial point clouds often lack sufficient detail and accuracy, constraining the performance of 3DGS and limiting its applicability to sparse-viewpoint scene reconstruction \cite{xiong2023sparsegs,chen2024optimizing,fan2024instantsplat}.

\begin{figure}[t]
   \centering
   \includegraphics[width=0.95\linewidth]{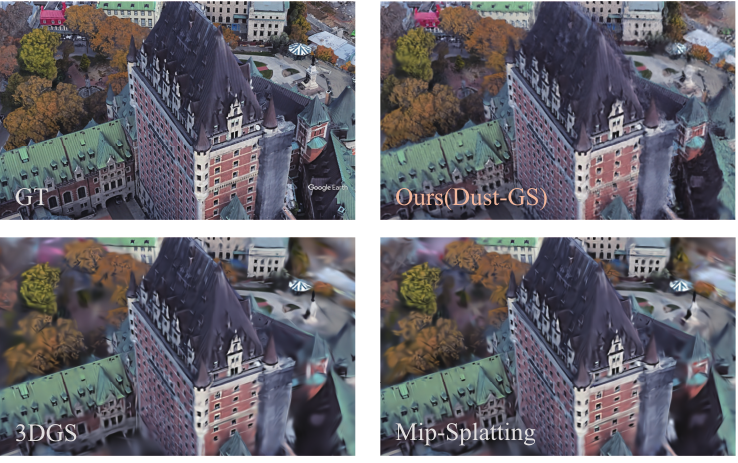}
   \caption{Novel View Synthesis Comparison. Comparison of 3DGS \cite{kerbl20233d} and Mip-Splatting \cite{yu2024mip} with our Dust-GS shows that Dust-GS outperforms the other methods in synthesizing close-up scenes.}
   \label{fig:mov}
 \end{figure}

\begin{figure*}[t]
\vspace{0.2cm}
   \centering
   \includegraphics[width=0.9\linewidth]{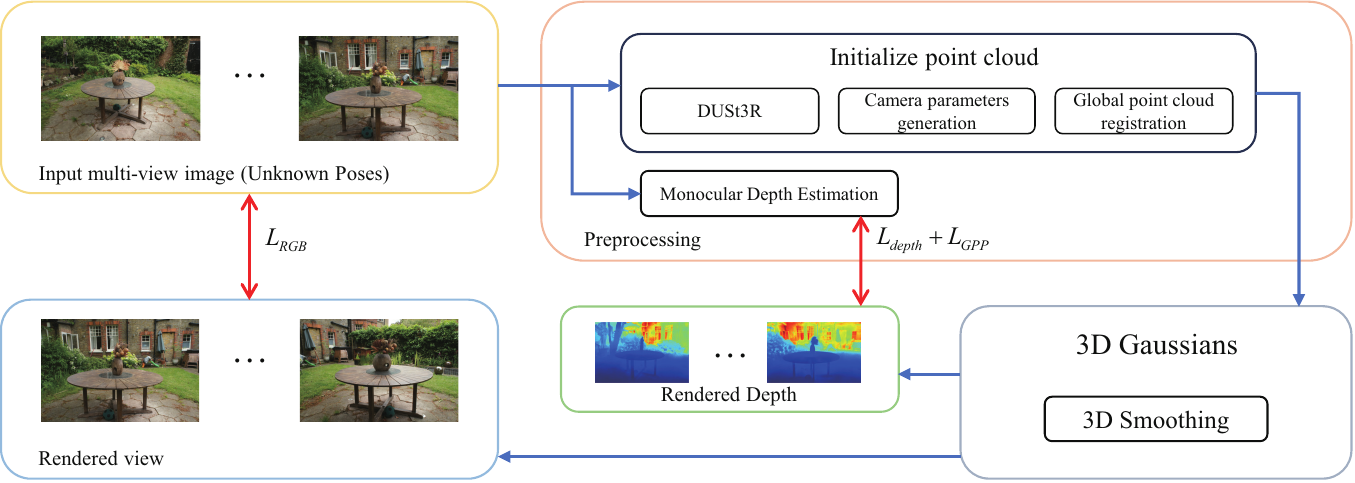}
   \caption{The framework estimates camera poses and registers point clouds using DUSt3R, initializes 3D Gaussian primitives, and optimizes them with RGB, depth, Gaussian Process Priors (GPP), and dynamic depth masks for improved scene reconstruction.}
   \label{fig:Pipeline}
 \end{figure*}

To address the limitations of 3DGS in sparse viewpoint conditions, we propose that the density of point clouds is crucial for high-quality scene reconstruction. We introduce Dust-GS, a novel framework designed to enhance point cloud initialization beyond traditional SfM methods. 
Dust-GS integrates a hybrid strategy that combines adaptive depth-based masking with point cloud optimization to generate denser and more accurate point clouds from sparse input data. This approach demonstrates that "dense point clouds matter" for achieving superior scene synthesis under limited viewpoints.
The comparison of Dust-GS with other methods is shown in Fig.\ref{fig:mov}.
To further enhance 3D reconstruction accuracy and stability in sparse data settings, Dust-GS introduces an adaptive depth processing technique and an improved point cloud initialization strategy that generates denser point clouds, overcoming the sparsity issues of SfM-based methods. Additionally, a dynamic depth masking mechanism is employed to filter high-frequency noise and artifacts from distant objects, retaining critical geometric information from mid- and near-range areas. This targeted depth management significantly improves scene reconstruction quality from sparse inputs, as demonstrated by our experimental results. 


In summary, our contributions are:
\begin{itemize}
    \item We propose a framework that optimizes Gaussian splatting for detailed 3D scene reconstruction using sparse input data, enhancing performance in few-shot conditions.
    \item A novel point cloud initialization strategy is introduced, reducing reliance on traditional SfM methods and generating denser, more accurate point clouds for improved scene synthesis.
    \item We develop a dynamic depth masking mechanism to refine the control over Gaussian primitives, effectively filtering noise and improving geometric fidelity in the reconstructed scenes.
\end{itemize}

\section{method}
The motivation of Dust-GS is to address the performance degradation of 3DGS under sparse viewpoint conditions. In particular, when input data is limited, point clouds of SfM  tend to be overly sparse, resulting in missing geometric details and inaccurate scene reconstruction. 
The entire framework is shown in the Fig.\ref{fig:Pipeline}.
Dust-GS introduces a new point cloud initialization strategy and an adaptive depth estimation mask to reduce reliance on dense input data, while maintaining high geometric consistency and view synthesis quality.

\subsection{3D Gaussian Splatting}
3D Gaussian Splatting (3DGS) is an explicit representation-based rendering method that uses Gaussian primitives  for representing 3D scenes. 3DGS explicitly defines the position $\mu\in \mathbb{R}^{3}$, color $c\in \mathbb{R}^{3}$, opacity $\alpha\in \mathbb{R}^{1}$, and covariance $\Sigma \in \mathbb{R}^{7}$ of each Gaussian primitive to perform rendering directly in 3D space:
\begin{equation}
     \mathcal{G}_n(p) = e^{-\frac{1}{2}(p-\mu_n)^T \Sigma^{-1}_n (p-\mu_n)}.
\end{equation}
To render the RGB image from a given viewpoint, the color $c$ is computed using spherical harmonic (SH) coefficients.
On the projected 2D plane, the color is determined via $C_(p) = \sum_{i = 1}^m c_i \sigma_i \prod_{j=1}^{i-1} (1 - \sigma_j)$.
$\sigma_i$ is determined by the covariance of the 2D Gaussian multiplied by the learned opacity $\alpha$ of each Gaussian ($\sigma_i = \alpha_i \mathcal{G}^{2D}_i(x)$).
In summary, 3DGS uses a set of 3D Gaussian primitives $\{\mathcal{G}_i|i = 1, 2, \ldots, n\} $ to represent a scene.

\subsection{Initialize point cloud}
3DGS faces the challenge of optimizing the initialization from either random point clouds or point clouds generated by SfM.
DUSt3R \cite{wang2024dust3r} takes two images as input and directly outputs per-pixel point maps and confidence maps.
For 3DGS initialization, it is necessary to obtain the intrinsic and extrinsic camera parameters corresponding to each input image.
The focal of each camera is calculated using the Weiszfeld algorithm \cite{plastria2011weiszfeld}:
\begin{equation}
    f^* = \arg\min_{f} \sum_{i=0}^{W} \sum_{j=0}^{H} \omega^{i,j} \left\| (i', j') - f  \frac{(P^{i,j}_0, P^{i,j}_1)}{P^{i,j}_2}  \right\|,
\end{equation}
where \( i' = i - \frac{W}{2} \) and \( j' = j - \frac{H}{2} \) denote the centered pixel indices; $P$ represents the pointmap.  Finally, the focal lengths from all training views are averaged to represent the camera focal length $\bar{f} = mean(f^*)$. 

To extend pair-wise aligned camera poses to globally aligned ones, we first construct a complete connectivity graph \( \mathcal{P}(\mathcal{V}, \mathcal{E}) \), where the vertices \( \mathcal{V} \) represent the \( N \) input views, and edges \( \mathcal{E} \) indicate pairs of images that share visual content. Given any image pair \( I_n \) and \( I_m \), we optimize the transformation matrix \( T_e \), scaling factor \( \sigma_e \), and globally aligned point map \( \tilde{P} \) to minimize the following objective:
\begin{equation}
    \tilde{P}^* = \arg \min_{\tilde{P}, T, \sigma} \sum_{e \in \mathcal{E}} \sum_{v \in e} \sum_{i=1}^{HW} \omega_{v,e}^i \left\| \tilde{P}_v^i - \sigma_e T_e P_{v,e}^i \right\|,
\end{equation}
where \( \omega_{v,e}^i \) represents confidence weights, and \( P_{v,e}^i \) are the point maps corresponding to view \( v \) in \( e \). To avoid trivial solutions such as \( \sigma_e = 0 \), DUSt3R enforce the constraint \( \prod_{e} \sigma_e = 1 \).

\begin{figure*}[t]
\vspace{0.2cm}
   \centering
   \includegraphics[width=0.95\linewidth]{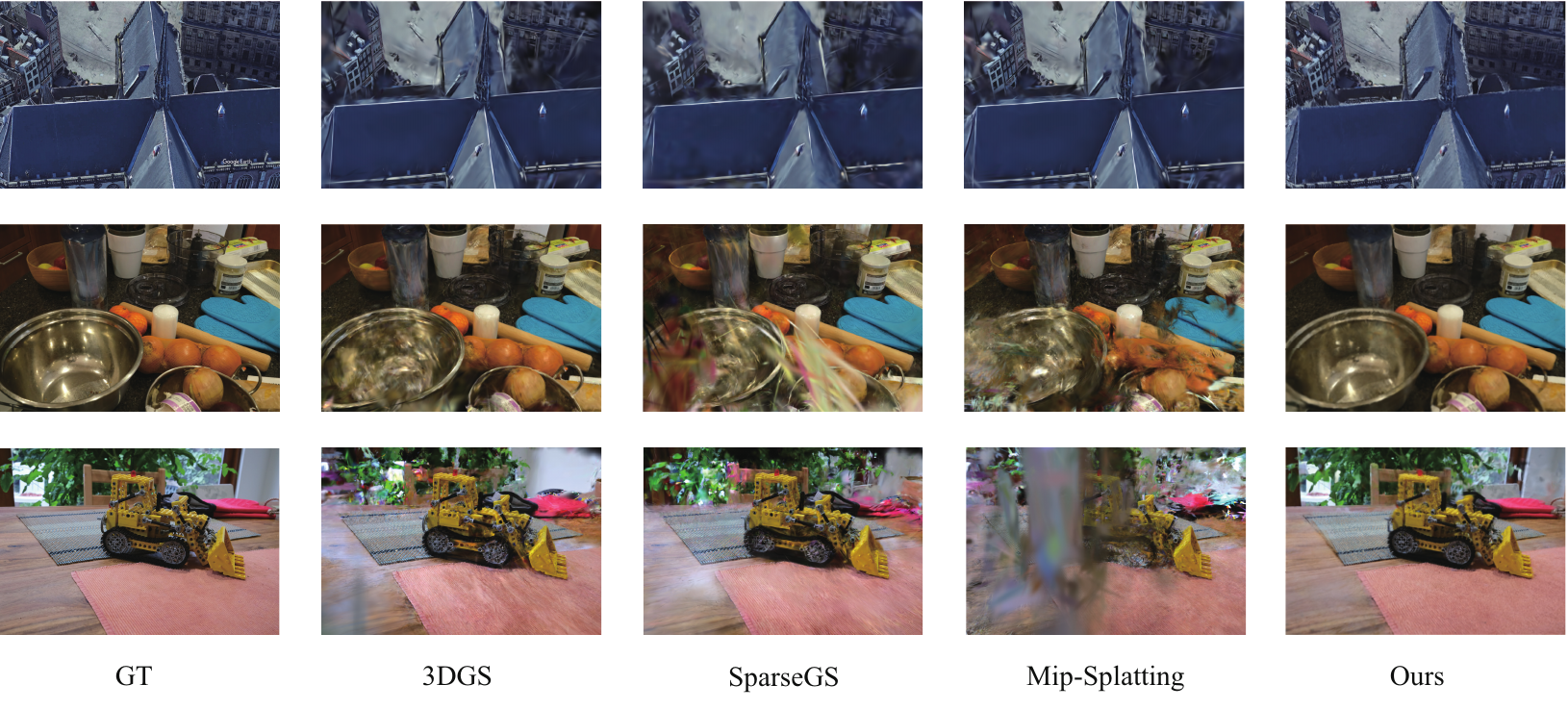}
   \caption{Qualitative results on the MipNeRF360 and BungeeNeRF datasets. Ours(Dust-GS) achieves superior geometric consistency and detail fidelity compared to 3DGS \cite{kerbl20233d}, SparseGS \cite{xiong2023sparsegs}, and Mip-Splatting \cite{yu2024mip}, with results that are closer to the ground truth (GT).}
   \label{fig:Qualitative}
 \end{figure*}

\subsection{Depth Correlation Loss}
Depth Correlation Loss is introduced to enforce consistency between predicted depth maps across multiple views, ensuring that corresponding pixels in different views maintain accurate depth relationships. 
The depth map for each view is obtained by accumulating the values of \(N\) ordered Gaussian primitives along the ray:
\begin{equation}
    D_r(p) = \sum_{i \in N} d_{i} \sigma_i \prod_{j=1}^{i-1}(1 - \sigma_j),
\end{equation}
where \(d_{i}\) represents the depth of the \(i\)-th Gaussian primitive. 

The Pearson Correlation Coefficient (PCC) is employed to measure the similarity between the predicted depth patch \( D \) and the ground truth depth patch \( \Hat{D}^{gt} \):
\begin{equation}
    \mathcal{L}_{\text{depth}} = \frac{1}{N} \sum_{i=1}^{N} \left( 1 - \text{PCC}(D_r, \Hat{D}_r^{gt}) \right),
\end{equation}
where the PCC is defined as:
\begin{equation}
    \text{PCC}(A, B) = \frac{\mathbb{E}[AB] - \mathbb{E}[A]\mathbb{E}[B]}{\sqrt{\mathbb{E}[A^2] - \mathbb{E}[A]^2} \sqrt{\mathbb{E}[B^2] - \mathbb{E}[B]^2}}.
\end{equation}

\subsection{Dynamic depth mask}
To mitigate noise and irrelevant details in the distant background, which can blur edges and reduce reconstruction quality, we introduce a dynamic depth mask. This mask suppresses high-frequency noise from distant objects and enhances geometric accuracy. The far-distance threshold \( s_f \) is dynamically calculated as:
\begin{equation}
    s_f = q_b + \left( \frac{\mu_D}{\sqrt{\mu_D^2 + \sigma_D^2}} \right) \times \Delta q,
\end{equation}
where \( \mu_D \) and \( \sigma_D \) are the mean and standard deviation of the depth map \( D \); \( q_b \) is the base quantile; \( \Delta q \) is the dynamic adjustment range. 

The masked depth map \( D_m = D \odot M \) is used for gradient operations, where $M  = Mask(D, s_f)$. The Gradient Profile Prior (GPP) \cite{sun2010gradient} further sharpens edges by aligning the gradient profiles of the rendered depth map $D_m$ with the target depth map \( \hat{D}_n \) after mask. The GPP loss is defined as:
\begin{equation}
    \mathcal{L}_{\text{GPP}}(D) = \frac{1}{u_1 - u_0} \int_{u_0}^{u_1} \| \nabla \hat{D}^{gt}_m(u) - \nabla D_m(u) \| \, du,
\end{equation}
where \( \nabla D\) represents the gradient field of the depth map. 

\subsection{Loss Function}
To optimize the 3D Gaussian representation, we formulate an optimization objective that incorporates multiple loss terms. The final loss function used to refine the 3D Gaussian parameters is expressed as follows: 
\begin{equation}
\begin{aligned}
    \mathcal{L}_{final} & = \mathcal{L}_{\text{RGB}}(I(\theta),\hat{I}) + \lambda_{\text{depth}} \mathcal{L}_{\text{depth}}(D(\theta),\hat{D})\\
    & + \lambda_{\text{GPP}}\mathcal{L}_{\text{GPP}}(D(\theta),\hat{D},M,I,\hat{I})
\end{aligned}
\end{equation}
where $I$, $D$ and the RGB images and depth maps rendered by the 3D Gaussian primitives; $\hat{I}$, $\hat{D}$ represent the reference RGB images and depth maps.

\section{EXPERIMENTAL RESULTS}
\subsection{Datasets and Evaluation Metrics}
We evaluate our method on two datasets: MipNeRF360 \cite{barron2022mip} and BungeeNeRF \cite{xiangli2022bungeenerf}. MipNeRF360 dataset provides complex, large-scale scenes with a wide range of viewpoints. BungeeNeRF dataset consists of multi-scale urban images captured using Google Earth Studio, designed to simulate data acquired by cameras at different altitudes.
For both datasets, we use three standard evaluation metrics to assess the quality of the reconstructed 3D scenes: Peak Signal-to-Noise Ratio (PSNR), Structural Similarity Index (SSIM) \cite{wang2004image}, and Learned Perceptual Image Patch Similarity (LPIPS) \cite{zhang2018unreasonable}. PSNR evaluates the overall image quality, SSIM measures structural similarity between images, and LPIPS captures perceptual differences between generated and ground truth images.

\begin{table}[t]
\centering
\caption{Quantitative Evaluation. Comparison of methods on Mip-NeRF360 and BungeeNeRF datasets. On the datasets with 8 input views, Dust-GS outperforms other methods.}
\resizebox{0.45\textwidth}{!}{%
\begin{tabular}{cc|ccc|ccc}
\hline
 & \multirow{2}{*}{\textbf{Method}} & \multicolumn{3}{c|}{\textbf{Mip-NeRF360}} & \multicolumn{3}{c}{\textbf{BungeeNeRF}}  \\ \cline{3-8} 
& & \textbf{PSNR}$\uparrow$ & \textbf{SSIM}$\uparrow$ & \textbf{LPIPS}$\downarrow$ & \textbf{PSNR}$\uparrow$ & \textbf{SSIM}$\uparrow$ & \textbf{LPIPS}$\downarrow$ \\ \hline
& \textbf{3DGS} & 9.89 & 0.141 & 0.592 & 17.22 & 0.431 & 0.591  \\ 
& \textbf{SparseGS} & 10.55 & 0.189 & 0.598 & 18.09 & 0.499 & 0.564  \\ 
& \textbf{InstantSplat} & 12.43 & 0.203 & 0.577 & 18.59 & 0.571 & 0.340  \\ 
& \textbf{Mip-Splatting} & 10.43 & 0.148 & 0.605 & 17.74 & 0.444 & 0.581  \\ 
& \textbf{Ours} & 12.58 & 0.210 & 0.583 & 18.60 & 0.567 & 0.346  \\ \hline
\end{tabular}%
}
\label{table:Quantitative}
\end{table}

\subsection{Implementation Details}
Our method is implemented using PyTorch \cite{paszke2019pytorch} and is trained and evaluated on a single $\text{NVIDIA}^{\text{TM}}$ RTX 4090 GPU. For optimization, we use the Adam optimizer, and the model is trained for 1000 iterations. 
We incorporate a 3D smoothing filter \cite{yu2024mip} to constrain the size of the 3D primitives.
During training, we apply a combination of depth-based loss terms, including the depth correlation loss and Gradient Profile Prior (GPP) loss, to ensure geometric consistency and sharpness in the final rendered images.
For all training and testing scenes, the same loss function and hyperparameters were used to optimize the 3D Gaussian primitives, with the model trained for 1K iterations.

\subsection{Qualitative and Quantitative Evaluation}
The quantitative results in the Table.\ref{table:Quantitative} show that Dust-GS outperforms other methods across both datasets. In terms of PSNR and SSIM, Dust-GS demonstrates superior image reconstruction quality compared to other methods, indicating its stronger ability to handle geometric structures and image details. Furthermore, Dust-GS significantly outperforms traditional 3DGS and SparseGS in LPIPS, suggesting notable improvements in visual quality and perceptual accuracy. While other methods approach Dust-GS in certain metrics, there remains a performance gap in overall results. In summary, by incorporating improved strategies, Dust-GS achieves higher geometric fidelity and visual consistency in 3D scene reconstruction under sparse viewpoints.

As shown in Fig.\ref{fig:Qualitative}, the qualitative analysis results demonstrate significant differences in visual reconstruction quality across the various methods. Compared to the ground truth (GT), the traditional 3DGS method struggles with detail restoration and geometric consistency, especially in complex scenes where images exhibit blurring and distortion. SparseGS incorporates depth information into 3DGS, leading to some improvements in synthesis quality. Although Mip-Splatting restricts the size of Gaussian primitives, image distortion remains prominent, particularly in handling object edges and complex textures, resulting in weak scene consistency.
In contrast, our method consistently delivers better visual quality across all test scenes, with sharper details and improved geometric consistency. Dust-GS notably excels in scenes with reflective objects and complex lighting conditions, reducing blurring and distortion, and achieving higher reconstruction accuracy compared to other methods.

\subsection{Ablations and Analysis}
As shown in Table.\ref{table:Abl}, we conducted ablation studies to evaluate the impact of key components in our method. 
The ablation study reveals that removing any key component of the framework results in a clear degradation in performance. In particular, the Depth Correlation Loss is essential for maintaining consistency in depth across multiple views, as its removal led to a noticeable decline in reconstruction accuracy. Similarly, the absence of the 3D smoothing filter resulted in reduced geometric consistency and surface smoothness, showing its importance in mitigating high-frequency noise and producing more visually coherent surfaces. The 3D smoothing step plays a critical role in enhancing surface detail, making the reconstructed geometry appear smoother and more natural. Furthermore, the dynamic depth mask proved crucial in filtering out irrelevant distant information and sharpening the geometric representation of the scene. Overall, the full model with all components included performs best, demonstrating that each element contributes significantly to the improvement of reconstruction quality.

\begin{table}[t]
\centering
\caption{Ablation studies on underwater scenes. Comparisons on the BungeeNeRF dataset with 8 input views indicate that the model with all modules performs best, with each module contributing to the overall performance.}
\begin{tabularx}{0.45\textwidth}{l*{3}{>{\centering\arraybackslash}X}}
\toprule
 & \textbf{PSNR}$\uparrow$ & \textbf{SSIM}$\uparrow$ & \textbf{LPIPS}$\downarrow$\\
\midrule
w/o Depth Correlation Loss & 12.57 & 0.239 & 0.587\\
w/o 3D smoothing & 13.80 & 0.238 & 0.561\\
w/o Dynamic depth mask & 13.75 & 0.236 & 0.563\\
All & 13.85 & 0.242 & 0.557\\
\bottomrule
\end{tabularx}

\label{table:Abl}
\end{table}

\section{Conclusion}
In this paper, we presented a novel framework for sparse viewpoint 3D scene reconstruction that effectively addresses the challenges associated with data sparsity in 3DGS-based methods. By introducing a dynamic depth mask and novel point clouds initialization strategy, we significantly improved the accuracy and robustness of scene reconstruction in scenarios with limited input images. Our method successfully enhances geometric consistency, suppresses noise, and preserves fine details, even under challenging sparse-view conditions. Experimental results on multiple benchmark datasets demonstrate the superiority of our approach. Our framework exhibits robust performance across a variety of challenging environments, especially in scenarios where acquiring dense multi-view data is difficult, such as in robotics, autonomous driving, and virtual reality applications.


\bibliographystyle{IEEEtran}
\bibliography{refs}

\begin{thebibliography}{10}
\providecommand{\url}[1]{#1}
\csname url@samestyle\endcsname
\providecommand{\newblock}{\relax}
\providecommand{\bibinfo}[2]{#2}
\providecommand{\BIBentrySTDinterwordspacing}{\spaceskip=0pt\relax}
\providecommand{\BIBentryALTinterwordstretchfactor}{4}
\providecommand{\BIBentryALTinterwordspacing}{\spaceskip=\fontdimen2\font plus
\BIBentryALTinterwordstretchfactor\fontdimen3\font minus \fontdimen4\font\relax}
\providecommand{\BIBforeignlanguage}[2]{{%
\expandafter\ifx\csname l@#1\endcsname\relax
\typeout{** WARNING: IEEEtran.bst: No hyphenation pattern has been}%
\typeout{** loaded for the language `#1'. Using the pattern for}%
\typeout{** the default language instead.}%
\else
\language=\csname l@#1\endcsname
\fi
#2}}
\providecommand{\BIBdecl}{\relax}
\BIBdecl

\bibitem{tonderski2024neurad}
A.~Tonderski, C.~Lindstr{\"o}m, G.~Hess, W.~Ljungbergh, L.~Svensson, and C.~Petersson, ``Neurad: Neural rendering for autonomous driving,'' in \emph{Proceedings of the IEEE/CVF Conference on Computer Vision and Pattern Recognition}, 2024, pp. 14\,895--14\,904.

\bibitem{pan2023uniocc}
M.~Pan, L.~Liu, J.~Liu, P.~Huang, L.~Wang, S.~Zhang, S.~Xu, Z.~Lai, and K.~Yang, ``Uniocc: Unifying vision-centric 3d occupancy prediction with geometric and semantic rendering,'' \emph{arXiv preprint arXiv:2306.09117}, 2023.

\bibitem{zhou2024drivinggaussian}
X.~Zhou, Z.~Lin, X.~Shan, Y.~Wang, D.~Sun, and M.-H. Yang, ``Drivinggaussian: Composite gaussian splatting for surrounding dynamic autonomous driving scenes,'' in \emph{Proceedings of the IEEE/CVF Conference on Computer Vision and Pattern Recognition}, 2024, pp. 21\,634--21\,643.

\bibitem{martin2021nerf}
R.~Martin-Brualla, N.~Radwan, M.~S. Sajjadi, J.~T. Barron, A.~Dosovitskiy, and D.~Duckworth, ``Nerf in the wild: Neural radiance fields for unconstrained photo collections,'' in \emph{Proceedings of the IEEE/CVF conference on computer vision and pattern recognition}, 2021, pp. 7210--7219.

\bibitem{mildenhall2021nerf}
B.~Mildenhall, P.~P. Srinivasan, M.~Tancik, J.~T. Barron, R.~Ramamoorthi, and R.~Ng, ``Nerf: Representing scenes as neural radiance fields for view synthesis,'' \emph{Communications of the ACM}, vol.~65, no.~1, pp. 99--106, 2021.

\bibitem{wang2023sparsenerf}
G.~Wang, Z.~Chen, C.~C. Loy, and Z.~Liu, ``Sparsenerf: Distilling depth ranking for few-shot novel view synthesis,'' in \emph{Proceedings of the IEEE/CVF International Conference on Computer Vision}, 2023, pp. 9065--9076.

\bibitem{truong2023sparf}
P.~Truong, M.-J. Rakotosaona, F.~Manhardt, and F.~Tombari, ``Sparf: Neural radiance fields from sparse and noisy poses,'' in \emph{Proceedings of the IEEE/CVF Conference on Computer Vision and Pattern Recognition}, 2023, pp. 4190--4200.

\bibitem{kerbl20233d}
B.~Kerbl, G.~Kopanas, T.~Leimk{\"u}hler, and G.~Drettakis, ``3d gaussian splatting for real-time radiance field rendering.'' \emph{ACM Trans. Graph.}, vol.~42, no.~4, pp. 139--1, 2023.

\bibitem{li2024dngaussian}
J.~Li, J.~Zhang, X.~Bai, J.~Zheng, X.~Ning, J.~Zhou, and L.~Gu, ``Dngaussian: Optimizing sparse-view 3d gaussian radiance fields with global-local depth normalization,'' in \emph{Proceedings of the IEEE/CVF Conference on Computer Vision and Pattern Recognition}, 2024, pp. 20\,775--20\,785.

\bibitem{lu2024scaffold}
T.~Lu, M.~Yu, L.~Xu, Y.~Xiangli, L.~Wang, D.~Lin, and B.~Dai, ``Scaffold-gs: Structured 3d gaussians for view-adaptive rendering,'' in \emph{Proceedings of the IEEE/CVF Conference on Computer Vision and Pattern Recognition}, 2024, pp. 20\,654--20\,664.

\bibitem{zhang2024pixel}
Z.~Zhang, W.~Hu, Y.~Lao, T.~He, and H.~Zhao, ``Pixel-gs: Density control with pixel-aware gradient for 3d gaussian splatting,'' \emph{arXiv preprint arXiv:2403.15530}, 2024.

\bibitem{yu2024mip}
Z.~Yu, A.~Chen, B.~Huang, T.~Sattler, and A.~Geiger, ``Mip-splatting: Alias-free 3d gaussian splatting,'' in \emph{Proceedings of the IEEE/CVF Conference on Computer Vision and Pattern Recognition}, 2024, pp. 19\,447--19\,456.

\bibitem{schoenberger2016mvs}
J.~L. Sch\"{o}nberger, E.~Zheng, M.~Pollefeys, and J.-M. Frahm, ``Pixelwise view selection for unstructured multi-view stereo,'' in \emph{European Conference on Computer Vision (ECCV)}, 2016.

\bibitem{schonberger2016structure}
J.~L. Schonberger and J.-M. Frahm, ``Structure-from-motion revisited,'' in \emph{Proceedings of the IEEE conference on computer vision and pattern recognition}, 2016, pp. 4104--4113.

\bibitem{xiong2023sparsegs}
H.~Xiong, S.~Muttukuru, R.~Upadhyay, P.~Chari, and A.~Kadambi, ``Sparsegs: Real-time 360 $\{$$\backslash$deg$\}$ sparse view synthesis using gaussian splatting,'' \emph{arXiv preprint arXiv:2312.00206}, 2023.

\bibitem{chen2024optimizing}
S.~Chen, J.~Zhou, and L.~Li, ``Optimizing 3d gaussian splatting for sparse viewpoint scene reconstruction,'' \emph{arXiv preprint arXiv:2409.03213}, 2024.

\bibitem{fan2024instantsplat}
Z.~Fan, W.~Cong, K.~Wen, K.~Wang, J.~Zhang, X.~Ding, D.~Xu, B.~Ivanovic, M.~Pavone, G.~Pavlakos \emph{et~al.}, ``Instantsplat: Unbounded sparse-view pose-free gaussian splatting in 40 seconds,'' \emph{arXiv preprint arXiv:2403.20309}, 2024.

\bibitem{wang2024dust3r}
S.~Wang, V.~Leroy, Y.~Cabon, B.~Chidlovskii, and J.~Revaud, ``Dust3r: Geometric 3d vision made easy,'' in \emph{Proceedings of the IEEE/CVF Conference on Computer Vision and Pattern Recognition}, 2024, pp. 20\,697--20\,709.

\bibitem{plastria2011weiszfeld}
F.~Plastria, ``The weiszfeld algorithm: proof, amendments, and extensions,'' \emph{Foundations of location analysis}, pp. 357--389, 2011.

\bibitem{sun2010gradient}
J.~Sun, Z.~Xu, and H.-Y. Shum, ``Gradient profile prior and its applications in image super-resolution and enhancement,'' \emph{IEEE Transactions on Image Processing}, vol.~20, no.~6, pp. 1529--1542, 2010.

\bibitem{barron2022mip}
J.~T. Barron, B.~Mildenhall, D.~Verbin, P.~P. Srinivasan, and P.~Hedman, ``Mip-nerf 360: Unbounded anti-aliased neural radiance fields,'' in \emph{Proceedings of the IEEE/CVF conference on computer vision and pattern recognition}, 2022, pp. 5470--5479.

\bibitem{xiangli2022bungeenerf}
Y.~Xiangli, L.~Xu, X.~Pan, N.~Zhao, A.~Rao, C.~Theobalt, B.~Dai, and D.~Lin, ``Bungeenerf: Progressive neural radiance field for extreme multi-scale scene rendering,'' in \emph{European conference on computer vision}.\hskip 1em plus 0.5em minus 0.4em\relax Springer, 2022, pp. 106--122.

\bibitem{wang2004image}
Z.~Wang, A.~C. Bovik, H.~R. Sheikh, and E.~P. Simoncelli, ``Image quality assessment: from error visibility to structural similarity,'' \emph{IEEE transactions on image processing}, vol.~13, no.~4, pp. 600--612, 2004.

\bibitem{zhang2018unreasonable}
R.~Zhang, P.~Isola, A.~A. Efros, E.~Shechtman, and O.~Wang, ``The unreasonable effectiveness of deep features as a perceptual metric,'' in \emph{Proceedings of the IEEE conference on computer vision and pattern recognition}, 2018, pp. 586--595.

\bibitem{paszke2019pytorch}
A.~Paszke, S.~Gross, F.~Massa, A.~Lerer, J.~Bradbury, G.~Chanan, T.~Killeen, Z.~Lin, N.~Gimelshein, L.~Antiga \emph{et~al.}, ``Pytorch: An imperative style, high-performance deep learning library,'' \emph{Advances in neural information processing systems}, vol.~32, 2019.

\end{thebibliography}

\end{document}